\documentclass[11pt]{article}
\usepackage{amssymb}
\usepackage[final]{acl}

\usepackage{times}
\usepackage{latexsym}
\usepackage{tikz}
\usepackage{amssymb}
\usepackage{pifont}
\usepackage{subcaption}
\usepackage{amsmath}
\usepackage{multirow}
\usepackage{booktabs}
\usepackage{bigstrut}
\usepackage{arydshln}  
\usepackage{amsmath}
\usepackage[most]{tcolorbox}
\usepackage[table,xcdraw]{xcolor}
\usepackage{makecell}
\tcbuselibrary{skins}
\tcbuselibrary{breakable}
\usepackage{colortbl}
\usepackage{array}

\usepackage[T1]{fontenc}

\usepackage[utf8]{inputenc}

\usepackage{microtype}

\usepackage{inconsolata}

\usepackage{graphicx}

\title{From Coarse to Fine: Self-Adaptive Hierarchical Planning for LLM Agents}

\author{
 \textbf{Haoran Tan\textsuperscript{1}\footnotemark[2]\footnotemark[3]},
 \textbf{Zeyu Zhang\textsuperscript{1}\footnotemark[2]\footnotemark[3]},
 \textbf{Chen Ma\textsuperscript{1}\footnotemark[2]\footnotemark[3]},
 \textbf{Tianze Liu\textsuperscript{2}},
\\
 \textbf{Quanyu Dai\textsuperscript{2}},
 \textbf{Xu Chen\textsuperscript{1}\footnotemark[1]\footnotemark[2]\footnotemark[3]},
\\
\\
 \textsuperscript{1}Gaoling School of Artificial, Renmin University of China, Beijing, China,\\
 \textsuperscript{2}Huawei Noah's Ark Lab
\\
\texttt{\{tanhaoran1321, xu.chen\}@ruc.edu.cn}
}

\begin{document}
\maketitle
\footnotetext[1]{Corresponding authors.}
\footnotetext[2]{Beijing Key Laboratory of Research on Large Models and Intelligent Governance}
\footnotetext[3]{Engineering Research Center of Next-Generation Intelligent Search and Recommendation, MOE}

\begin{abstract}
Large language model-based agents have recently emerged as powerful approaches for solving dynamic and multi-step tasks.
Most existing agents employ planning mechanisms to guide long-term actions in dynamic environments.
However, current planning approaches face a fundamental limitation that they operate at a fixed granularity level.
Specifically, they either provide excessive detail for simple tasks or insufficient detail for complex ones, failing to achieve an optimal balance between simplicity and complexity.
Drawing inspiration from the principle of \textit{progressive refinement} in cognitive science, we propose \textbf{AdaPlan-H}, a self-adaptive hierarchical planning mechanism that mimics human planning strategies. Our method initiates with a coarse-grained macro plan and progressively refines it based on task complexity. It generates self-adaptive hierarchical plans tailored to the varying difficulty levels of different tasks, which can be optimized by imitation learning and capability enhancement.
Experimental results demonstrate that our method significantly improves task execution success rates while mitigating overplanning at the planning level, providing a flexible and efficient solution for multi-step complex decision-making tasks.
To contribute to the community, our code and data will be made publicly available at \url{https://github.com/import-myself/AHP}.
\end{abstract}

\section{Introduction}

Large language model (LLM) based agents are increasingly important for solving complex tasks that require multi-step decision-making~\citep{wang2024survey}.
The planning capability of an agent enables it to better understand the long-term dependencies within these tasks and to take actions in the correct sequence in dynamic environments~\citep{cao2025large}.
Most previous works attempt to design LLM-based planners that can generate explicit task plannings to guide the actor's execution process~\citep{xiong2025mpo, erdogan2025plan, wang2023plan}. Some of them leverage searching methods for task decomposition~\citep{prasad2023adapt} or explore within the planning space~\cite{yao2023tree, hu2024treeplannerefficientcloselooptask}.
Besides, some works also try to empower LLM with implicit planning capabilities by finetuning-based methods that update LLM's parameters using expert trajectories or feedback~\citep{zeng2023agenttuning}, which aims to directly increase the probability of LLM-based agents taking correct actions given contexts~\citep{cao2025large}. 

However, although explicit planning can improve the consistency of agents' actions, most existing methods directly generate plans at a fixed granularity.
These single-level methods fail to address the need for less detail in simple tasks and full expansion in complex tasks. 
As a result, too coarse planning granularity can lead to insufficient effectiveness, while too fine granularity will reduce the agent's task completion efficiency.
Some decomposition-based methods implement task divisions, but they either rigidly break down initial plans without granularity adaptation, or require iterative trial-and-error to decide decomposition~\citep{prasad2023adapt}.
\textbf{Therefore, our target is to adaptively generate plans with proper granularity based on different tasks}.

\begin{figure*}[t]
\includegraphics[width=\textwidth]{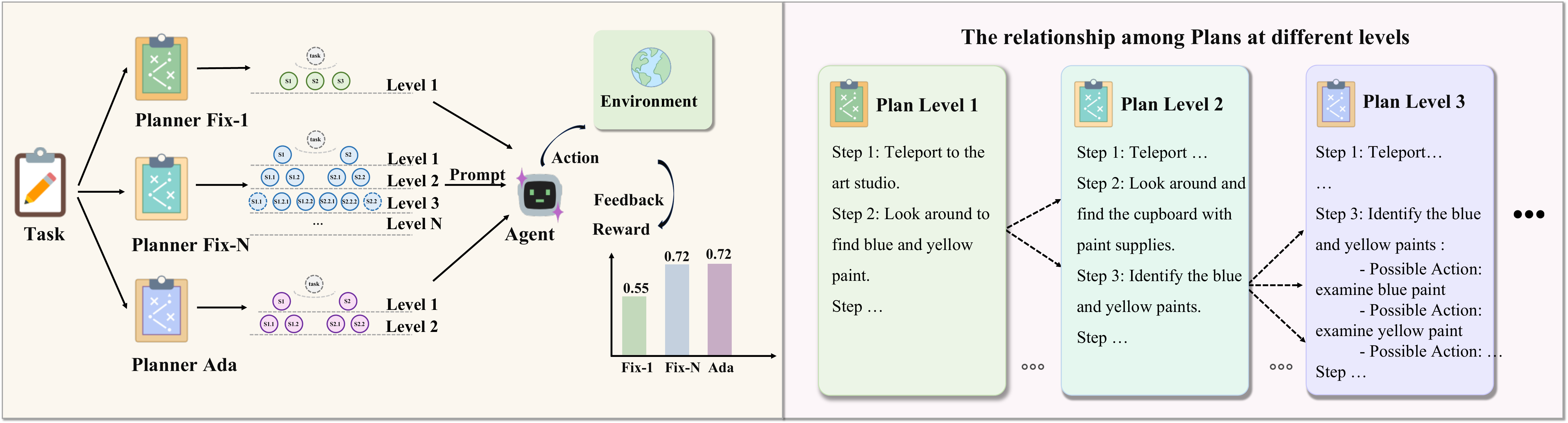}
  \caption{The left part illustrates how hierarchical plans at different levels assist the agent in interacting with the environment and receiving reward feedback. The right part illustrates the primary distinction between different levels lies in the degree of detail, with a clear inheritance relationship between them.}
  \vspace{-0.5cm}
  \label{fig:motivation}    
\end{figure*}

In cognitive science, researchers find that natural human planning typically starts with formulating a high-level coarse macro goal, which can be gradually refined into more specific plans until concrete steps are determined.
This process is also known as \textit{progressive refinement}, and has been validated in both simple tasks (\textit{e.g.,} cooking) and complex tasks (\textit{e.g.,} medical diagnosis)~\citep{zacks2001perceiving}.
This progressive refinement approach can effectively reduce the action search space and simplify complex tasks~\citep{correa2025exploring}.

Inspired by the \textit{progressive refinement} theory, we propose a self-adaptive hierarchical planning framework to address our core problem, called \textbf{AdaPlan-H}.
Our method empowers the agent to start with a coarse-grained global plan and progressively expand the details based on task complexity, forming a hierarchical planning structure. It mimics human planning strategies by starting from a macro plan and refining it step by step.
Besides, we design imitation learning and capability enhancement procedures to optimize our framework.
In addition, we conduct experiments to verify that our framework can generate plans with enough details for task effectiveness and improve task efficiency by avoiding overplanning at the planning level.
To benefit the research community, we release our code and data at \url{https://github.com/import-myself/AHP}.

In summary, our contributions are as follows:\\
$\bullet$ We propose a self-adaptive hierarchical planning framework for LLM-based agents to generate explicit plans with proper granularity based on different tasks, called \textbf{AdaPlan-H}.\\
$\bullet$ We design imitation learning and capability enhancement procedures to optimize our framework for self-adaptive hierarchical planning generation.\\
$\bullet$ We conduct experiments to verify that our method can significantly improve task efficiency and success rates, along with extensive experiments.

\section{Preliminary}

We model the tasks with environmental feedback for an agent as a Partially Observable Markov Decision Process (POMDP)~\citep{cao2025large}, defined as a tuple $(U, S, A, O, T, R)$, where $U$ is the task instruction space, $S$ is the state space, $A$ is the action space, $O$ is the observation space, $T: S \times A \to S$ is the state transition function, and $R: S \times A \to \mathbb{R}$ is the reward function. Notably, for LLM-based agents, $U$, $A$, and $O$ are spaces of natural language.

In this paper, our agent focuses on two components, including the planner and the actor \citep{erdogan2025plan}.
When interacting with the environment, the planner can generate a plan based on the task instruction and interaction context.
Then, the actor will follow this plan to guide the generation of sequential actions, interact with the environment to complete the task, and receive feedback.
Specifically, given a task instruction $u \in U$, the planner of an LLM-based agent can generate a plan \( p = f_p(u, \theta_p) \), where \( \theta_p \) represents the parameters of the planner. The actor module then generates the first action $a_1 \sim \pi_{\theta_a}(\cdot | u, p)$ based on its policy $\pi_{\theta_a}$, and executing this action causes a transition in the latent state space $s_t \in S$. The environment then provides feedback $o_t \in O$ as a new observation. At time step $t$, the agent generates the next action $a_{t+1} \sim \pi_{\theta_a}(\cdot | u, a_1, o_1, \dots, o_{t-1}, a_t, p)$ based on the task instruction $u$, the observation $o_t$, and the plan $p$. This interaction process repeats until either the task is successfully completed or the maximum allowed number of steps is reached. 
The task  trajectory can be denoted as
\[
e = (u, a_1, o_1, \dots, o_{n-1}, a_n).
\]
Finally, the task's trajectory feedback, which is the final reward $r$, is provided by the environment.

\section{Method}
Inspired by the theory of \textit{progressive refinement} in cognitive science~\citep{zacks2001perceiving}, we define the hierarchical planning and design a self-adaptive method to generate it based on different tasks. 
Building on this, we develop a two-stage optimization framework, including imitation learning and capability enhancement.
After initially endowing the planner with the ability to generate adaptive hierarchical plans through imitation learning, 
we further enhance the accuracy of the hierarchy and the quality of the planning.

\subsection{Self-Adaptive Hierarchical Planning}
Our study focuses on the process that the planner of an LLM-based agent generates a plan 
\( p = f_p(u, \theta_p) \)
, where \(\theta_p\) represents the
parameters of the planner.
Cognitive science theories on hierarchical planning and \textit{progressive refinement} suggest that human planning typically starts with a high-level coarse macro goal and then gradually refines the goal into more specific plans~\citep{zacks2001perceiving, correa2025exploring}. 
Inspired by this theory, we model the process of generating the plan $p$ as a hierarchical generation process. Unlike previous methods where the planner directly generates a final plan $p$, our hierarchical planning first generates the highest-level plan based on the instruction.
Then, it generates the $i$-th level plan based on the task instruction and previous planning levels.
Therefore, the final plan can be represented as
\begin{align*}
    p &= p_m \oplus p_{m-1} \oplus \dots \oplus p_1, \\
    p_i &= f_p(u, p_{i-1}, p_{i-2},...p_1, \theta_p),\\
    p_1 &= f_p(u, \theta_p),
\end{align*}
where $m$ represents the number of levels in the hierarchical plan and $\oplus$ indicates concatenation. In our work, $m$ is a self-adaptive variable that changes according to the task complexity and difficulty.

As illustrated in Figure~\ref{fig:motivation}, each level of the hierarchical plan corresponds to a complete plan for accomplishing the task, with the key difference being the level of detail. That is, a plan at a higher level is further refined in the next level, adding additional conditions or details. 
During the agent's actor execution phase, plans are presented hierarchically. The high-level plans serve as global guidance, while the low-level plans provide more detailed instructions. Through this hierarchical structure, the agent not only receives clearer guidance for each task but also ensures alignment with the overall task through the global plan at higher levels. Since the details involved in explicit plans may change with the environment, the lower-level plans are more likely to undergo changes.

\begin{figure*}[t]
  \includegraphics[width=\textwidth]{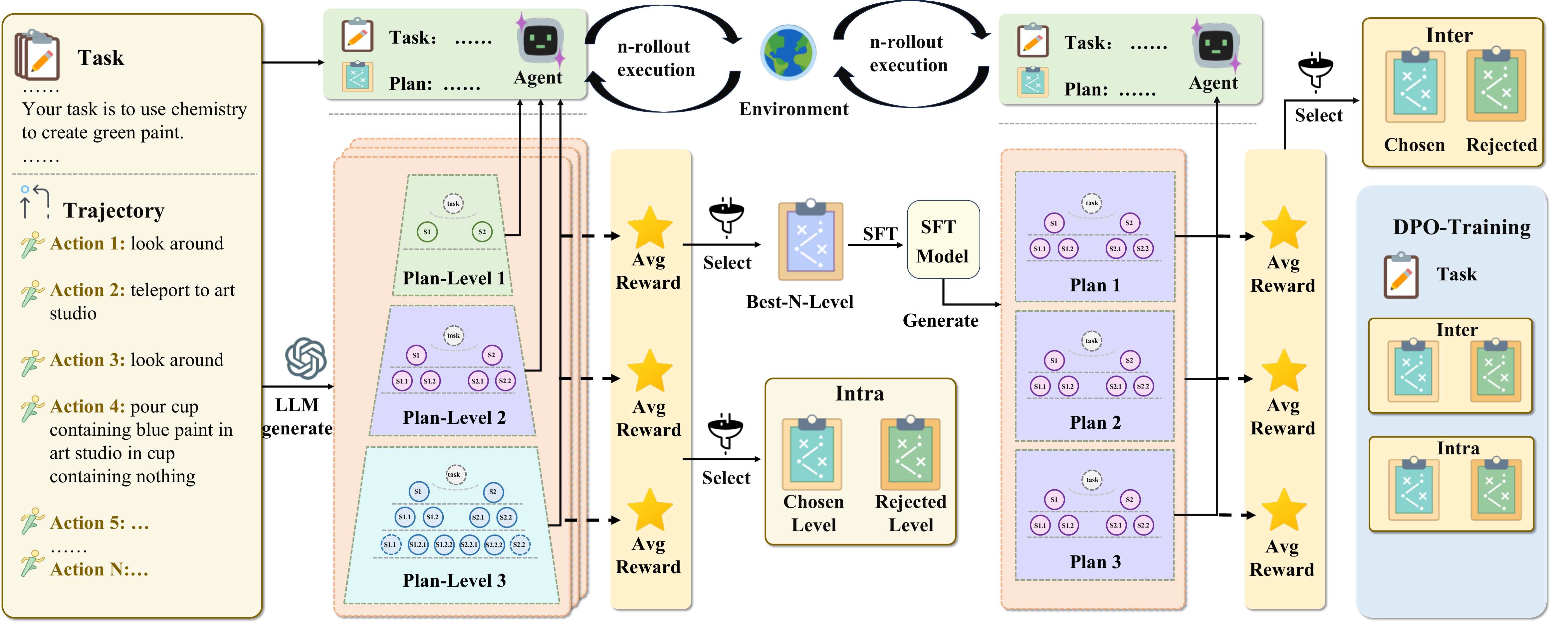}
  \caption{The overall architecture of AdaPlan-H with two-stage optimization. First of all, we construct the optimal level plan and perform SFT initialization. After that, the hierarchical preference pairs and quality preference pairs are used for DPO training to achieve self-adaptive hierarchical explicit planning.}
  \vspace{-0.4cm}
  \label{fig:framework}
\end{figure*}

\subsection{Imitation Learning}

\subsubsection{Hierarchical Plan Generation}

To enable the planner to learn adaptive hierarchical planning based on task difficulty and complexity, we use supervised fine-tuning in the first stage to initialize the model. Therefore, we need a training set of tasks paired with the corresponding best hierarchical plans. However, existing datasets often only provide correct trajectories for completing tasks, which can be used for implicit plan training but are not suitable for training explicit guidance. 
Thus, we utilize the large model GPT-4o~\citep{achiam2023gpt} and the correct task completion trajectories to generate $N$ hierarchical plans for each task instruction. 
It is important to note that we cannot directly obtain the optimal number of hierarchical levels for each task. Therefore, we first generate a plan with a fixed maximum number of levels $M$.
Relevant prompts can be found in Appendix~\ref{sec:prompt}.

\subsubsection{Hierarchical Plan Selection}

Existing datasets do not provide labels for the best level of hierarchical plan. Besides, the difficulty of tasks is subjective, making it difficult to construct supervised signals for hierarchies through manual or LLM-based annotation.
Therefore, we adopt the Monte Carlo method from \citet{xiong2025mpo} to evaluate each level of each hierarchical plan through exploration. We decompose each hierarchical plan into $M$ hierarchical plans, where we denote \(p_i\) as the first $i$ levels of $p$. 
It will yield a total of $N \times M$ plans for each task. Each hierarchical plan \(p_m^n\), represents a plan constructed by extracting the first $m$ levels from the plan $n$, where \(m \in \{1,...,M\}, n \in \{ 1,...,N\} \).
These hierarchical plans are then inserted into the agent's prompt respectively, and the agent performs the corresponding task $K$ times for each hierarchical plan. 
Based on a hierarchical plan \(p_m^n\), the trajectory generated by the agent can be represented as
\begin{equation*}
\begin{split}
    \{e^{(i)}_{n,m}|i =1,\dots, K\}
    \sim \pi_{\theta_{a}}(e|u, p_{m}^n).
\end{split}
\end{equation*}
The environment then provides feedback in the form of a reward $r(u, e^{(i)}_{n,m})$. Thus, we compute the average reward for each hierarchical plan \(p_m^n\) as:
\begin{equation*}
   Q(p_m^n) = \frac{1}{K} \sum_{i=1}^{K} r(u, e^{(i)}_{n,m} ). 
\end{equation*}

For evaluation in Monte Carlo method, we use Llama-3.1-8B-Instruct~\citep{dubey2024llama} as the actor in our agent framework, as this model strikes a balance in instruction following and task completion, making the plans it generates highly adaptable and universal.
By comparing the average rewards, we can select the best plan \(p_{best}\) with the highest average reward and the lowest level, that is,
\begin{equation*}
    p_{\text{best}} = \arg \min_m \left( \max_n Q(p_m^n) \right).
\end{equation*}

\subsubsection{Supervised Fine-tuning Initialization}
\label{sec:sft_init}
This selected best plan is then used as the supervised output for the corresponding task instruction, and the chosen number of best level $m$ serves as the hierarchical supervision signal. Our Supervised Fine-Tuning (SFT) dataset can be represented as
\begin{equation*}
    D_s = \{(u, p_{\text{best}})^{(i)}\}_{i=1}^{|D_s|}.
\end{equation*}
Then, we can use this dataset to perform SFT on our planner, enabling it to initialize the ability of self-adaptive hierarchical planning with the loss
\begin{equation*}
   L_{SFT} = -\mathbb{E}_{(u, p_{\text{best}}) \sim D_s} \Big[ \log \pi_{\theta_p}(p_{\text{best}}|u)\Big].
\end{equation*}

\subsection{Capability Enhancement}

\subsubsection{Level Preference Data Construction}
Since we now have the hierarchical supervision signal for tasks, we can construct a dataset emphasizing the level preferences of plans, denoted as the intra-dataset. It can be represented as:
\begin{equation*}
\begin{split}
        D_{d_{intra}} = \{(u, p_m^n, p_{m'}^{n'})^{(i)}\}_{i=1}^{|D_{d_{intra}}|}, \\
        m' \neq m, n' \neq n,
\end{split}
\end{equation*}
where $p_m^n$ is chosen and $p_{m'}^{n'}$ is rejected. Here, $m$ and $n$ correspond to the values of $p_{\text{best}}$ in stage 1. The different values of $m$ represent preferences for different task hierarchies, while different values of $n$ are used to ensure that they are not derived from the same plan generated by GPT-4o~\citep{achiam2023gpt} before. 
Since we plan to use Direct Preference Optimization (DPO), the shared prefix tokens are excluded from the loss computation ~\citep{pal2024smaug}, we must choose hierarchical plans extracted from different plans generated before.

\subsubsection{Quality Preference Data Construction}

After the SFT initialization, we aim to further enhance the planner’s ability in both the accuracy of hierarchical planning levels and the quality of the generated plans. To achieve this, we intend to use DPO as the optimization algorithm, and we need to construct quality preference data. We generate multiple plans using the SFT-initialized model. Assuming the model has almost acquired adaptive hierarchical planning ability, we can select the most common number of levels $m$ as the ground truth hierarchies for each task and discard other plans whose level count does not equal $m$. 
Then, we set the task to yield $N$ plans with exactly $m$ levels, \textit{i.e.,} \(p_n, n\in\{1,...,N\}\), and we also use Monte Carlo method in Section~\ref{sec:sft_init} to assess these plans' quality by their average reward in exploration.
After that, we insert these plans into the actor's prompt and execute the task $K$ times. For each task, the trajectory for the plan $p_n$ is
\begin{equation*}
    \{e^{(i)}_n|i =1,\dots,K\} \sim \pi_{\theta_a}(e|u,p^n).
\end{equation*}
The average reward for each plan is
\begin{equation*}
    Q(p^n) = \frac{1}{K} \sum_{i=1}^{K} r(u, e^{(i)}_n ).
\end{equation*}
Then, we construct a preference pair dataset emphasizing plan quality, denoted as the inter-dataset:
\begin{equation*}
    D_{d_{inter}} = \{(u, p^n, p^{n'})^{(i)}\}_{i=1}^{|D_{d_{inter}}|}, n' \neq n,
\end{equation*}
where $p^n$ is chosen and $p^{n'}$ is rejected.
It is important to note that these plans have a fixed number of levels $m$, which is previously determined.

\subsubsection{DPO Training}

We combine the intra-dataset and the inter-dataset to create the training dataset for DPO:
\begin{equation*}
    D_d = D_{d_{intra}} + D_{d_{inter}}.
\end{equation*}
DPO optimizes the model by increasing the likelihood of selected plans relative to rejected ones. However, it is sensitive to sequence length. Even with balanced data distributions, its sensitivity to length can cause the model to output longer sequences~\citep{liu2024length}. 
Since the intra-dataset emphasizes level preferences, which are represented as the length of plan's overall tokens, directly using DPO loss would not effectively train the planner model. 
To address this, we modify the DPO loss by adding an SFT loss term:
\begin{align*}
    & L_{DPO(\pi_\theta;\pi_{ref})} = - \mathbb{E}_{(u, p_{\text{c}}, p_{\text{r}}) \sim D_d} 
    \Big[ \log \sigma \\
    &\left( \beta \log \frac{\pi_{\theta_p}(p_{\text{c}}|u)}{\pi_{ref}(p_{\text{c}}|u)}
        - \beta \log \frac{\pi_{\theta_p}(p_{\text{r}}|u)}{\pi_{ref}(p_{\text{r}}|u)} \right) \Big]\\
&        - \gamma \mathbb{E}_{(u, p_{\text{c}}) \sim D_d} \Big[ \log \pi_{\theta_p}(p_{\text{c}}|u) \Big],
\end{align*}
where $p_c$ is the chosen plan, $p_r$ is the rejected plan, and $\gamma$ represents the coefficient multiplied by the SFT loss. To prevent the SFT loss from dominating, we set the hyper-parameter $\gamma \in[0,1]$.

Through DPO training, we enable the planner to generate hierarchical plans with more accurate levels and higher quality. It can better guide the agent's actor to complete tasks and avoid over in the planning process.

\begin{table*}[htbp]
  \centering
  \caption{Performance of different methods on two datasets under various actor backbones. The \textit{Seen} refers to the validation set, and \textit{Unseen} refers to the test set. The \textit{w/o plan} indicates whether explicit planning generated by the planner is used as guidance. The bolded values represent the optimal results.}
  \label{tab:main results}%
  \vspace{-0.1cm}
  \resizebox{\textwidth}{!}
{
\begin{tabular}{ccccccccc}
\hline
\hline
\bf{Datasets} & \bf{Methods} & \bf{w/o Plan} & \bf{Llama-3.1-8B} & \bf{Qwen2.5-7B} & \bf{Qwen3-8B} & \bf{Chatglm-9B} &  \bf{GPT-4o-mini} \bigstrut\\
\hline
\multirow{4}[4]{*}{ALFWorld (Seen)} & ReAct & $\times$ & 0.2500 & 0.7173 & 0.7286 & 0.5286 &  0.4357 \bigstrut[t]  \\
& Base Planner & $\checkmark$ & 0.2214 & 0.5643 & 0.5714 & 0.4429 &  0.4071 \bigstrut[t] \\
& MPO & $\checkmark$ & 0.5000 & 0.8286 & 0.7687 & 0.6643 &  0.7643 \bigstrut[t] \\
& \cellcolor{gray!20}AdaPlan-H & \cellcolor{gray!20}$\checkmark$ & \cellcolor{gray!20}\textbf{0.5429} & \cellcolor{gray!20}\bf0.8500 & \cellcolor{gray!20}\bf0.8357 & \cellcolor{gray!20}\bf0.86766 &  \cellcolor{gray!20}\bf0.8429 \bigstrut[t]\\
\hline
\multirow{4}[4]{*}{ALFWorld (Unseen)} & ReAct & $\times$  & 0.1866 & 0.7537 & 0.7164 & 0.5448 &  0.4030 \bigstrut[t]  \\
& Base Planner & \checkmark & 0.2090 & 0.5448 & 0.5299 & 0.5522 & 0.3881 \bigstrut[t] \\
& MPO & $\checkmark$ & 0.5224 & 0.8284 & 0.7388 & 0.6940 &  0.7836 \bigstrut[t] \\
& \cellcolor{gray!20}AdaPlan-H& \cellcolor{gray!20}\checkmark & \cellcolor{gray!20}\bf0.5896 & \cellcolor{gray!20}\bf0.8881 & \cellcolor{gray!20}\bf0.7985 & \cellcolor{gray!20}\bf0.7230 &  \cellcolor{gray!20}\bf0.8284 \bigstrut[t]\\
\hline
\multirow{4}[4]{*}{ScienceWorld (Seen)} & ReAct & $\times$ & 0.4367 & 0.3103 & 0.5764 & 0.2923 &  0.5423 \bigstrut[t]  \\
& Base Planner & $\checkmark$ & 0.4817 & 0.4246 & 0.5798 & 0.3308 &  0.5884 \bigstrut[t] \\
& MPO & $\checkmark$ & 0.5724 & 0.4303 & 0.6608 & 0.3618 &  \bf0.6527 \bigstrut[t] \\
& \cellcolor{gray!20}AdaPlan-H& \cellcolor{gray!20}$\checkmark$ & \cellcolor{gray!20}\bf0.5876 & \cellcolor{gray!20}\bf0.4573 & \cellcolor{gray!20}\bf0.7349 & \cellcolor{gray!20}\bf0.4127 &  \cellcolor{gray!20}0.6454 \bigstrut[t]\\
\hline
\multirow{4}[4]{*}{ScienceWorld (Unseen)} & ReAct & $\times$ & 0.4233 & 0.2978 & 0.5087 & 0.3618 &  0.5031 \bigstrut[t]  \\
& Base Planner & $\checkmark$ & 0.4417 & 0.3844 & 0.5640 & 0.3245 &  0.5382 \bigstrut[t] \\
& MPO & $\checkmark$  & 0.5458 & 0.437 & 0.6444 & \bf0.3719 &  0.6044 \bigstrut[t] \\
& \cellcolor{gray!20}AdaPlan-H & \cellcolor{gray!20}$\checkmark$ & \cellcolor{gray!20}\bf0.5531 & \cellcolor{gray!20}\bf0.4412 & \cellcolor{gray!20}\bf0.6766 & \cellcolor{gray!20}0.3693 &  \cellcolor{gray!20}\bf0.6155 \bigstrut[t]\\
\hline
\hline
\end{tabular}%
}
\vspace{-0.4cm}
\end{table*}%

\section{Experiments}
\subsection{Experimental Settings}
\textbf{Datasets.} We focus on multi-step decision-making scenarios with long-term dependencies. To this end, we conduct experiments on two representative agent datasets exhibiting these characteristics: ALFWorld~\citep{shridhar2020alfworld} for embodied household tasks and ScienceWorld~\citep{wang2022scienceworld} for text-based science experiments. Both ALFWorld and ScienceWorld require agents to output multi-step sequential actions in textual form. Upon task completion or reaching the maximum number of steps, the environment provides a reward signal: ALFWorld offers a binary reward indicating task success or not, while ScienceWorld provides a dense reward ranging from 0 to 1. 
These two datasets both include validation sets that are in-distribution (seen) and test sets that are both out-of-distribution (unseen) relative to the training set.
Following the setting in MPO~\citep{xiong2025mpo}, we treat both validation and test sets as test sets to evaluate the effectiveness and generalization of our method. Additionally, we follow ETO~\citep{song2024trial} by leveraging their open-sourced correct trajectories on the training sets of these datasets to generate initial hierarchical plans for our training data in training stage 1. And the detailed statistics of these datasets can be found in Appendix~\ref{sec: data}.

\textbf{Planner Methods.} We employ Llama-3.1-8B as the backbone model for our self-adaptive hierarchical planner. For comparison, we evaluate against several commonly used explicit guidance-based planning methods: (1) ReAct~\citep{yao2023react}: The agent reasons before taking an action during decision-making; (2) Base Planner: A standard prompting-based planning approach without training the planner, utilizing in-context learning with examples to generate plans. We use GPT-4o for its plan generation. (3) MPO~\citep{xiong2025mpo}: After initializing the planner via SFT on meta-plans generated by LLMs, MPO samples generations from the SFT model and applies the same Monte Carlo method as ours to construct preference pairs for DPO. From our perspective, MPO can be viewed as the method that only performs a single level of planning for all tasks. 

\textbf{Actor Models.} To validate the effectiveness of self-adaptive hierarchical planning across different actor models, we experiment with a variety of models as the actor: Llama-3.1-8B-Instruct~\citep{dubey2024llama}, Qwen2.5-7B-Instruct~\citep{qwen2.5}, Qwen3-8B~\citep{qwen3technicalreport}, GLM-4-9B-Chat~\citep{glm2024chatglm}, and GPT-4o-mini~\citep{achiam2023gpt}. Notably, the model used to construct hierarchical supervision signals in the first training stage is Llama-3.1-8B-Instruct. Thus, our actor models are ideally comparable in capability. Otherwise, the constructed hierarchies may introduce redundancy and bias for other models. To ensure the stability of the results, we set the generation parameter temperature of the actor model to 0.

\textbf{Implementation Details.} In the first stage of our training, we set the maximum number of levels for the initial hierarchical plans $M$ to 3 on both the ALFWorld and ScienceWorld datasets. For each task instruction, we sampled $N = 5$ plans with 3 levels, resulting in 15 hierarchical plans for each task. In the second stage, we set the temperature to 0.7 for the SFT-trained model to generate 2 alternative plans for constructing quality preference pairs. For the training process, we used \texttt{llama\_factory}~\citep{zheng2024llamafactory} to perform SFT and DPO training. For the generation process, we employed \texttt{vllm}~\citep{kwon2023efficient} to accelerate the generation process. Our training is all conducted using LoRA~\citep{hu2022lora}, with the LoRA parameters uniformly set to a rank of 16 and alpha of 32. In the first stage of training, the learning rate is set to $5 \times 10^{-5}$, and the model is trained for 3 epochs. In the second stage of DPO training, $\beta$ is set to 0.1, $\gamma$ is set to 1, and the learning rate remains $5 \times 10^{-5}$, with training for only 1 epoch.

\subsection{Main Results}
As shown in Table~\ref{tab:main results}, across all tasks and regardless of the paired actor model, our planner helps the agent achieve substantial improvements in average reward compared to no planning or prompt-based planning methods. In most tasks, our method also outperforms MPO. Although in a few tasks, the self-adaptive hierarchical planning method does not outperform MPO, our method still achieves comparable results. This gap may be due to a bias in our training process: in the first stage, we use Llama-3.1-8B as the actor for the optimal level of hierarchical plan selection, and in the second stage, the same model is also used for plan quality evaluation. This introduces discrepancies and bias with the actual downstream actor models, as different architectures or model sizes may require distinct prompt optimizations. The optimal value of levels and the most effective plan for Llama-3.1-8B may not necessarily apply to other actor models. In later experiments Section~\ref{sec:backbone}, we demonstrate that maintaining consistency between the actor model used for plan evaluation and the one used for testing leads to greater improvements. Overall, the plans generated by our planner achieved better results in assisting different actor models in completing tasks compared to other methods. This demonstrates the effectiveness of our approach and its generalization for models with similar capabilities.

\subsection{Ablation Study}
We conduct ablation experiments on each aspect of our method's two training stages, including SFT in stage 1, and two types of DPO data for DPO training in stage 2. 
Evaluations are performed on the seen datasets of ALFWorld and ScienceWorld, using Llama-3.1-8B as the actor model. Since these are in-distribution with the training set, facilitating a clearer assessment of each component's effectiveness.  
As shown in Table~\ref{tab:ablation}, w/o $\text{dpo}_{inter}$ means DPO training without the quality preference-inter dataset, w/o $\text{dpo}_{intra}$ means DPO training without the level preference-intra dataset, w/o dpo means no DPO training, and w/o sft means no SFT training.
We can find that, after SFT, the model's ability to generate adaptive hierarchical plans has already improved task accuracy. Further enhancement of the planning capability was achieved after the integration of the two types of data in DPO, and both preferences for the levels and for the quality of the hierarchical plan are indispensable. 

\renewcommand{\arraystretch}{1.1}
\begin{table}[t]
  \centering
  \caption{Results of the ablation study on each part of AdaPlan-H's training framework.}
  \vspace{-0.1cm}
  \resizebox{\linewidth}{!}{
    \begin{tabular}{>{\centering\arraybackslash}p{2.2cm}>{\centering\arraybackslash}p{2.9cm}>{\centering\arraybackslash}p{3.4cm}}
\hline
\hline
\bf{Methods} & \bf{ALFWorld (Seen)} & \bf{ScienceWorld (Seen)}\\
\hline
w/o dpo$_{inter}$ &  0.4610 & 0.5529\\
w/o dpo$_{intra}$ &  0.4863 & 0.4762\\
w/o sft & 0.2553 & 0.4825\\
w/o dpo & 0.4881 & 0.5103\\
\rowcolor{gray!20}AdaPlan-H & \textbf{0.5429} & \textbf{0.5876} \\
\hline
\hline
\end{tabular}%
}
  \label{tab:ablation}
  \vspace{-0.2cm}
\end{table}%

\subsection{Comparison of Fixed-Level Plans}

To further explore the planning performance with different fixed levels, we group the hierarchical plans in the first training stage of AdaPlan-H according to their planning levels.
Then, we conduct the complete two-stage training and test on the two datasets.
To enhance clarity in presentation, we abbreviate ALFWorld as ALF, ScienceWorld as SCI, and denote Seen and Unseen as S and U/S, respectively.
According to the results in Table~\ref{tab:fix}, we find that as the number of levels increases, the task completion accuracy tends to improve. However, an excessive number of levels could lead to a decline in performance. This validates the necessity of designing our method as an adaptive hierarchical approach, which aims to minimize the number of levels as much as possible while ensuring task accuracy, in order to avoid inefficiencies.

\begin{table}[t]
  \centering
  \caption{The performance of hierarchical plans with different fixed-levels on two datasets. }
  \vspace{-0.1cm}
  \resizebox{\linewidth}{!}{
    \begin{tabular}{>{\centering\arraybackslash}p{1.2cm}>{\centering\arraybackslash}p{1.6cm}>{\centering\arraybackslash}p{1.7cm}>{\centering\arraybackslash}p{1.6cm}>{\centering\arraybackslash}p{1.7cm}}
\hline
\hline
\bf{Methods} & \bf{ALF (S)} & \bf{ALF (U/S)} & \bf{SCI (S)} & \bf{SCI (U/S)} \\
\hline
Fix-1 & 0.2571 & 0.2388 & 0.5332 & 0.4961 \\
Fix-2 & \textbf{0.5286} & 0.3806 & 0.5014 & 0.5258 \\
Fix-3 & 0.5000 & \textbf{0.5075} & \textbf{0.6261} & \textbf{0.5601} \\
\hline
\hline
\end{tabular}%
}
  \label{tab:fix}
  \vspace{-0.4cm}
\end{table}%

\subsection{Influence of Different Backbones}
\label{sec:backbone}

We further evaluate the influence of different backbones in our planner and actor.
According to the results shown in Table~\ref{tab:planner and acot}, we find that the choice of the planner's backbone model does influence the actor's performance. However, it is not necessarily optimal for the planner and actor to use the same backbone. Overall, using Llama-3.1-8B as the planner performs better than Qwen2.5-7B. However, in AdaPlan-H, we can still choose the planner's backbone model based on resource requirements. It is worth noting that, in this experiment, when one model is selected as the actor, this model is also used in the Monte Carlo method of plan evaluation. It leads to significantly better results in Qwen2.5-7B than training data obtained with Llama-3.1-8B in the Monte Carlo method, as shown in the main Table~\ref{tab:main results}. This suggests that we should strive to maintain consistency between the actor used in the actual test and the actor in the Monte Carlo method.

\begin{table}[t]
  \centering
  \caption{The influence of the different backbones in our planner and actor. \textit{Llama} indicates Llama-3.1-8B-Instruct, and \textit{Qwen} indicates Qwen2.5-7B-Instruct.}
  \vspace{-0.1cm}
  \resizebox{\linewidth}{!}{
    \begin{tabular}{>{\centering\arraybackslash}p{1cm}>{\centering\arraybackslash}p{1cm}>{\centering\arraybackslash}p{1.35cm}>{\centering\arraybackslash}p{1.7cm}>{\centering\arraybackslash}p{1.2cm}>{\centering\arraybackslash}p{1.6cm}}
\hline
\hline
\bf{Planner} & \bf{Actor} & \bf{ALF (S)} & \bf{ALF (U/S)} & \bf{SCI (S)} & \bf{SCI (U/S)} \\
\hline
Llama & Llama & 0.5429 & 0.5896 & 0.5876 & 0.5531\bigstrut[t]\\
Qwen &  Llama & 0.3929 & 0.3955 & 0.5600 & \textbf{0.5656} \\
Llama & Qwen  & \textbf{0.7500} & \textbf{0.7687} & \textbf{0.5925} & 0.5242 \\
Qwen & Qwen   & 0.6643 & 0.7338 & 0.5227 & 0.5504 \\
\hline
\hline
\end{tabular}%
}
  \label{tab:planner and acot}
  \vspace{-0.2cm}
\end{table}%

\subsection{Necessity of Maintaining Hierarchical Structure of Plan in Prompts}

We further explore whether our obtained plans are supposed to maintain a hierarchical structure in prompts or just keep the last level.
Therefore, we design \textit{Hierarchical} that indicates maintaining the hierarchical structure, while the \textit{Last-Level} indicates using only the last level.
As shown in Table~\ref{tab:specific}, the results indicate that maintaining the full hierarchical structure significantly improved task accuracy over only using the last level. This validates our previous hypothesis that the high-level plan in the hierarchical plan could maintain a broad alignment with the overall goal, while the low-level plan provides more detailed guidance. The hierarchical plans assist in task completion effectively.

\begin{table}[t]
  \centering
  \caption{The impact of maintaining the hierarchical structure of the plan as a prompt input on the task.}
  \vspace{-0.1cm}
  \resizebox{\linewidth}{!}{
    \begin{tabular}{ccccc}
\hline
\hline
\bf{Method} & \bf{ALF (S)} & \bf{ALF (U/S)} & \bf{SCI (S)} & \bf{SCI (U/S)} \\
\hline
Last-level & \bf{0.5571} & 0.5448 & 0.5454 & 0.5136 \bigstrut[t]\\
\rowcolor{gray!20}Hierarchical &  0.5425 & \bf{0.5896} & \bf{0.5876} & \bf{0.5531} \\
\hline
\hline
\end{tabular}%
}
  \label{tab:specific}
  \vspace{-0.5cm}
\end{table}%

\section{Related Work}
\textbf{LLM Agents.} As large language models have demonstrated significant potential in human intelligence~\citep{zhao2023survey}, research on LLM-based autonomous agents has gained increasing attention~\citep{wang2024survey}. An LLM-based agent typically includes: a profile module for managing the agent's personalized information and knowledge~\cite{tseng2024two}, a memory module for storing historical experiences accumulated during the agent's interaction with the environment~\cite{zhang2024survey, tan2025membench}, a planning module for generating execution strategies based on task requirements~\cite{cao2025large}, and an action module for performing specific operations according to the plan. By parsing natural language inputs and generating coherent responses, these agents can perform a variety of functions, such as answering questions, writing, generating code, and so on.

\textbf{Agent Optimization.} 
Early studies enhanced this capability through prompting techniques~\citep{yao2023react, shinn2023reflexion}. Some approaches like FireAct~\citep{chen2023fireact} and AgentTuning~\citep{zeng2023agenttuning} attempt supervised fine-tuning of agents using expert trajectories constructed from expert data of teacher agents (e.g. GPT-4). Recently, many works utilize reinforcement learning to optimize agents through interactions with the environment and reward signals, such as ETO~\cite{song2024trial} and IPR~\cite{xiong2024watch} which employ direct preference optimization (DPO) to refine trajectory generation, and works like GiGPO~\citep{feng2025group} that apply online RL algorithms such as GRPO~\cite{shao2024deepseekmath} for trajectory-level and action-level optimization. However, these training-based methods require full retraining of the models within the agent, limiting their applicability to open-source models and lacking modularity with other components.

\textbf{Agent Planning} Agent planning enables intelligent agents to autonomously choose actions to achieve goals in dynamic environments. 
Some works enhance planning by training decision-making through supervised fine-tuning with expert trajectories or reinforcement learning~\citep{zeng2023agenttuning, song2024trial, xiong2024watch, feng2025group}. Decomposition methods, like Adapt~\citep{prasad2023adapt}, split tasks into subgoals for iterative execution, while exploration-based approaches, such as ToT\citep{yao2023tree} that models the reasoning process as a tree for exploration to find optimal decisions. 
Other approaches improve planning with external modules. 
Traditional methods commonly utilize LLMs to translate tasks into Planning Domain Definition Language for planning solvers~\citep{liu2023llm+}, while recent works often model agent as separate planner and actor modules, using prompting or fine-tuning to enhance the planning capabilities of the planner and improve task success~\citep{wang2023plan, xiong2025mpo}.

In contrast to prior methods that employ LLMs as single-level planners~\cite{xiong2025mpo}, AdaPlan-H transforms the planner's output from single-level to hierarchical. Unlike decomposition-based hierarchies~\citep{prasad2023adapt}, AdaPlan-H avoids trial-and-error and provides explicit guidance, yielding clearer inheritance relationships between steps across levels, making it more suitable for multi-step tasks with long-range dependencies. The adaptive hierarchical nature of AdaPlan-H not only outperforms single-level planning in effectiveness but also mitigates token wastage from overplanning in the plan generation process.

\section{Conclusion}

In this paper, we propose \textbf{AdaPlan-H}, a self-adaptive hierarchical planning framework that generates plans with appropriate granularity for different tasks.
It successfully integrates the theory \textit{progressive refinement} of cognitive science into the process of hierarchical plan generation, and can significantly enhance agent performance in complex multi-step decision-making tasks. 
With the optimization of imitation learning and capability enhancement, our framework can dynamically adjust planning levels and granularity, overcoming the limitations of fixed-hierarchy methods and demonstrating flexibility across tasks of varying complexity. 
Experimental results confirm its significant advantages in planning efficiency and effectiveness. Our future work will focus on extending 
\textbf{AdaPlan-H} to broader task scenarios and optimizing its adaptability in multi-modal environments for improved agent planning capabilities.

\section*{Limitations}
We limited the maximum value of hierarchical levels to 3 in our experiment on two datasets, while the experimental results suggest that deeper levels could yield further improvements. 
In our main experiments, the training data for the planner model were obtained through exploration performed by the Llama-3.1-8B actor. However, as noted in Section~\ref{sec:backbone}, performance gains from planner-generated plans are more pronounced when the actor during exploration matches the one used in testing. 
In the experimental setup, we generated only two plans when constructing quality preference pairs, though larger values could be set. This also demonstrates that there is room for further improvement in the effectiveness of adaptive hierarchical plans.
\section*{Ethics Statement}
The data used in this paper comes from publicly available licensed datasets, used for research purposes under their respective licenses. This work complies with ACL ethical policies, and we declare no ethical issues.

\section*{Acknowledgments}
This work is supported in part by National Natural Science Foundation of China (No. 62422215 and No. 62472427), Huawei Innovation Research Programs, Major Innovation \& Planning Inter-disciplinary Platform for the "DoubleFirst Class" Initiative, Renmin University of China, Public
Computing Cloud, Renmin University of China, fund for building world-class universities (disci-plines) of Renmin University of China.

\clearpage

\bibliography{custom}

\begin{thebibliography}{35}
\providecommand{\natexlab}[1]{#1}

\bibitem[{Achiam et~al.(2023)Achiam, Adler, Agarwal, Ahmad, Akkaya, Aleman, Almeida, Altenschmidt, Altman, Anadkat et~al.}]{achiam2023gpt}
Josh Achiam, Steven Adler, Sandhini Agarwal, Lama Ahmad, Ilge Akkaya, Florencia~Leoni Aleman, Diogo Almeida, Janko Altenschmidt, Sam Altman, Shyamal Anadkat, and 1 others. 2023.
\newblock Gpt-4 technical report.
\newblock \emph{arXiv preprint arXiv:2303.08774}.

\bibitem[{Cao et~al.(2025)Cao, Men, Liu, Zhang, Li, Lin, Sui, Cao, Liu, and Zhao}]{cao2025large}
Pengfei Cao, Tianyi Men, Wencan Liu, Jingwen Zhang, Xuzhao Li, Xixun Lin, Dianbo Sui, Yanan Cao, Kang Liu, and Jun Zhao. 2025.
\newblock Large language models for planning: A comprehensive and systematic survey.
\newblock \emph{arXiv preprint arXiv:2505.19683}.

\bibitem[{Chen et~al.(2023)Chen, Shu, Shareghi, Collier, Narasimhan, and Yao}]{chen2023fireact}
Baian Chen, Chang Shu, Ehsan Shareghi, Nigel Collier, Karthik Narasimhan, and Shunyu Yao. 2023.
\newblock Fireact: Toward language agent fine-tuning.
\newblock \emph{arXiv preprint arXiv:2310.05915}.

\bibitem[{Correa et~al.(2025)Correa, Sanborn, Ho, Callaway, Daw, and Griffiths}]{correa2025exploring}
Carlos~G Correa, Sophia Sanborn, Mark~K Ho, Frederick Callaway, Nathaniel~D Daw, and Thomas~L Griffiths. 2025.
\newblock Exploring the hierarchical structure of human plans via program generation.
\newblock \emph{Cognition}, 255:105990.

\bibitem[{Dubey et~al.(2024)Dubey, Jauhri, Pandey, Kadian, Al-Dahle, Letman, Mathur, Schelten, Yang, Fan et~al.}]{dubey2024llama}
Abhimanyu Dubey, Abhinav Jauhri, Abhinav Pandey, Abhishek Kadian, Ahmad Al-Dahle, Aiesha Letman, Akhil Mathur, Alan Schelten, Amy Yang, Angela Fan, and 1 others. 2024.
\newblock The llama 3 herd of models.
\newblock \emph{arXiv e-prints}, pages arXiv--2407.

\bibitem[{Erdogan et~al.(2025)Erdogan, Lee, Kim, Moon, Furuta, Anumanchipalli, Keutzer, and Gholami}]{erdogan2025plan}
Lutfi~Eren Erdogan, Nicholas Lee, Sehoon Kim, Suhong Moon, Hiroki Furuta, Gopala Anumanchipalli, Kurt Keutzer, and Amir Gholami. 2025.
\newblock Plan-and-act: Improving planning of agents for long-horizon tasks.
\newblock \emph{arXiv preprint arXiv:2503.09572}.

\bibitem[{Feng et~al.(2025)Feng, Xue, Liu, and An}]{feng2025group}
Lang Feng, Zhenghai Xue, Tingcong Liu, and Bo~An. 2025.
\newblock Group-in-group policy optimization for llm agent training.
\newblock \emph{arXiv preprint arXiv:2505.10978}.

\bibitem[{GLM et~al.(2024)GLM, Zeng, Xu, Wang, Zhang, Yin, Rojas, Feng, Zhao, Lai, Yu, Wang, Sun, Zhang, Cheng, Gui, Tang, Zhang, Li, Zhao, Wu, Zhong, Liu, Huang, Zhang, Zheng, Lu, Duan, Zhang, Cao, Yang, Tam, Zhao, Liu, Xia, Zhang, Gu, Lv, Liu, Liu, Yang, Song, Zhang, An, Xu, Niu, Yang, Li, Bai, Dong, Qi, Wang, Yang, Du, Hou, and Wang}]{glm2024chatglm}
Team GLM, Aohan Zeng, Bin Xu, Bowen Wang, Chenhui Zhang, Da~Yin, Diego Rojas, Guanyu Feng, Hanlin Zhao, Hanyu Lai, Hao Yu, Hongning Wang, Jiadai Sun, Jiajie Zhang, Jiale Cheng, Jiayi Gui, Jie Tang, Jing Zhang, Juanzi Li, and 37 others. 2024.
\newblock \href {https://arxiv.org/abs/2406.12793} {Chatglm: A family of large language models from glm-130b to glm-4 all tools}.
\newblock \emph{Preprint}, arXiv:2406.12793.

\bibitem[{Hu et~al.(2022)Hu, Shen, Wallis, Allen-Zhu, Li, Wang, Wang, Chen et~al.}]{hu2022lora}
Edward~J Hu, Yelong Shen, Phillip Wallis, Zeyuan Allen-Zhu, Yuanzhi Li, Shean Wang, Lu~Wang, Weizhu Chen, and 1 others. 2022.
\newblock Lora: Low-rank adaptation of large language models.
\newblock \emph{ICLR}, 1(2):3.

\bibitem[{Hu et~al.(2024)Hu, Mu, Yu, Ding, Wu, Shao, Chen, Wang, Qiao, and Luo}]{hu2024treeplannerefficientcloselooptask}
Mengkang Hu, Yao Mu, Xinmiao Yu, Mingyu Ding, Shiguang Wu, Wenqi Shao, Qiguang Chen, Bin Wang, Yu~Qiao, and Ping Luo. 2024.
\newblock \href {https://arxiv.org/abs/2310.08582} {Tree-planner: Efficient close-loop task planning with large language models}.
\newblock \emph{Preprint}, arXiv:2310.08582.

\bibitem[{Kwon et~al.(2023)Kwon, Li, Zhuang, Sheng, Zheng, Yu, Gonzalez, Zhang, and Stoica}]{kwon2023efficient}
Woosuk Kwon, Zhuohan Li, Siyuan Zhuang, Ying Sheng, Lianmin Zheng, Cody~Hao Yu, Joseph Gonzalez, Hao Zhang, and Ion Stoica. 2023.
\newblock Efficient memory management for large language model serving with pagedattention.
\newblock In \emph{Proceedings of the 29th symposium on operating systems principles}, pages 611--626.

\bibitem[{Liu et~al.(2023)Liu, Jiang, Zhang, Liu, Zhang, Biswas, and Stone}]{liu2023llm+}
Bo~Liu, Yuqian Jiang, Xiaohan Zhang, Qiang Liu, Shiqi Zhang, Joydeep Biswas, and Peter Stone. 2023.
\newblock Llm+ p: Empowering large language models with optimal planning proficiency.
\newblock \emph{arXiv preprint arXiv:2304.11477}.

\bibitem[{Liu et~al.(2024)Liu, Bai, Han, Weng, Xu, Cao, Wang, and Cai}]{liu2024length}
Wei Liu, Yang Bai, Chengcheng Han, Rongxiang Weng, Jun Xu, Xuezhi Cao, Jingang Wang, and Xunliang Cai. 2024.
\newblock Length desensitization in direct preference optimization.
\newblock \emph{arXiv preprint arXiv:2409.06411}.

\bibitem[{Pal et~al.(2024)Pal, Karkhanis, Dooley, Roberts, Naidu, and White}]{pal2024smaug}
Arka Pal, Deep Karkhanis, Samuel Dooley, Manley Roberts, Siddartha Naidu, and Colin White. 2024.
\newblock Smaug: Fixing failure modes of preference optimisation with dpo-positive.
\newblock \emph{arXiv preprint arXiv:2402.13228}.

\bibitem[{Prasad et~al.(2023)Prasad, Koller, Hartmann, Clark, Sabharwal, Bansal, and Khot}]{prasad2023adapt}
Archiki Prasad, Alexander Koller, Mareike Hartmann, Peter Clark, Ashish Sabharwal, Mohit Bansal, and Tushar Khot. 2023.
\newblock Adapt: As-needed decomposition and planning with language models.
\newblock \emph{arXiv preprint arXiv:2311.05772}.

\bibitem[{Shao et~al.(2024)Shao, Wang, Zhu, Xu, Song, Bi, Zhang, Zhang, Li, Wu et~al.}]{shao2024deepseekmath}
Zhihong Shao, Peiyi Wang, Qihao Zhu, Runxin Xu, Junxiao Song, Xiao Bi, Haowei Zhang, Mingchuan Zhang, YK~Li, Yang Wu, and 1 others. 2024.
\newblock Deepseekmath: Pushing the limits of mathematical reasoning in open language models.
\newblock \emph{arXiv preprint arXiv:2402.03300}.

\bibitem[{Shinn et~al.(2023)Shinn, Cassano, Gopinath, Narasimhan, and Yao}]{shinn2023reflexion}
Noah Shinn, Federico Cassano, Ashwin Gopinath, Karthik Narasimhan, and Shunyu Yao. 2023.
\newblock Reflexion: Language agents with verbal reinforcement learning.
\newblock \emph{Advances in Neural Information Processing Systems}, 36:8634--8652.

\bibitem[{Shridhar et~al.(2020)Shridhar, Yuan, C{\^o}t{\'e}, Bisk, Trischler, and Hausknecht}]{shridhar2020alfworld}
Mohit Shridhar, Xingdi Yuan, Marc-Alexandre C{\^o}t{\'e}, Yonatan Bisk, Adam Trischler, and Matthew Hausknecht. 2020.
\newblock Alfworld: Aligning text and embodied environments for interactive learning.
\newblock \emph{arXiv preprint arXiv:2010.03768}.

\bibitem[{Song et~al.(2024)Song, Yin, Yue, Huang, Li, and Lin}]{song2024trial}
Yifan Song, Da~Yin, Xiang Yue, Jie Huang, Sujian Li, and Bill~Yuchen Lin. 2024.
\newblock Trial and error: Exploration-based trajectory optimization for llm agents.
\newblock \emph{arXiv preprint arXiv:2403.02502}.

\bibitem[{Tan et~al.(2025)Tan, Zhang, Ma, Chen, Dai, and Dong}]{tan2025membench}
Haoran Tan, Zeyu Zhang, Chen Ma, Xu~Chen, Quanyu Dai, and Zhenhua Dong. 2025.
\newblock Membench: Towards more comprehensive evaluation on the memory of llm-based agents.
\newblock \emph{arXiv preprint arXiv:2506.21605}.

\bibitem[{Team(2024)}]{qwen2.5}
Qwen Team. 2024.
\newblock \href {https://qwenlm.github.io/blog/qwen2.5/} {Qwen2.5: A party of foundation models}.

\bibitem[{Team(2025)}]{qwen3technicalreport}
Qwen Team. 2025.
\newblock \href {https://arxiv.org/abs/2505.09388} {Qwen3 technical report}.
\newblock \emph{Preprint}, arXiv:2505.09388.

\bibitem[{Tseng et~al.(2024)Tseng, Huang, Hsiao, Chen, Huang, Meng, and Chen}]{tseng2024two}
Yu-Min Tseng, Yu-Chao Huang, Teng-Yun Hsiao, Wei-Lin Chen, Chao-Wei Huang, Yu~Meng, and Yun-Nung Chen. 2024.
\newblock Two tales of persona in llms: A survey of role-playing and personalization.
\newblock \emph{arXiv preprint arXiv:2406.01171}.

\bibitem[{Wang et~al.(2024)Wang, Ma, Feng, Zhang, Yang, Zhang, Chen, Tang, Chen, Lin et~al.}]{wang2024survey}
Lei Wang, Chen Ma, Xueyang Feng, Zeyu Zhang, Hao Yang, Jingsen Zhang, Zhiyuan Chen, Jiakai Tang, Xu~Chen, Yankai Lin, and 1 others. 2024.
\newblock A survey on large language model based autonomous agents.
\newblock \emph{Frontiers of Computer Science}, 18(6):186345.

\bibitem[{Wang et~al.(2023)Wang, Xu, Lan, Hu, Lan, Lee, and Lim}]{wang2023plan}
Lei Wang, Wanyu Xu, Yihuai Lan, Zhiqiang Hu, Yunshi Lan, Roy Ka-Wei Lee, and Ee-Peng Lim. 2023.
\newblock Plan-and-solve prompting: Improving zero-shot chain-of-thought reasoning by large language models.
\newblock \emph{arXiv preprint arXiv:2305.04091}.

\bibitem[{Wang et~al.(2022)Wang, Jansen, C{\^o}t{\'e}, and Ammanabrolu}]{wang2022scienceworld}
Ruoyao Wang, Peter Jansen, Marc-Alexandre C{\^o}t{\'e}, and Prithviraj Ammanabrolu. 2022.
\newblock Scienceworld: Is your agent smarter than a 5th grader?
\newblock \emph{arXiv preprint arXiv:2203.07540}.

\bibitem[{Xiong et~al.(2025)Xiong, Song, Dong, Zhao, Song, Wang, and Li}]{xiong2025mpo}
Weimin Xiong, Yifan Song, Qingxiu Dong, Bingchan Zhao, Feifan Song, Xun Wang, and Sujian Li. 2025.
\newblock Mpo: Boosting llm agents with meta plan optimization.
\newblock \emph{arXiv preprint arXiv:2503.02682}.

\bibitem[{Xiong et~al.(2024)Xiong, Song, Zhao, Wu, Wang, Wang, Li, Peng, and Li}]{xiong2024watch}
Weimin Xiong, Yifan Song, Xiutian Zhao, Wenhao Wu, Xun Wang, Ke~Wang, Cheng Li, Wei Peng, and Sujian Li. 2024.
\newblock Watch every step! llm agent learning via iterative step-level process refinement.
\newblock \emph{arXiv preprint arXiv:2406.11176}.

\bibitem[{Yao et~al.(2023{\natexlab{a}})Yao, Yu, Zhao, Shafran, Griffiths, Cao, and Narasimhan}]{yao2023tree}
Shunyu Yao, Dian Yu, Jeffrey Zhao, Izhak Shafran, Tom Griffiths, Yuan Cao, and Karthik Narasimhan. 2023{\natexlab{a}}.
\newblock Tree of thoughts: Deliberate problem solving with large language models.
\newblock \emph{Advances in neural information processing systems}, 36:11809--11822.

\bibitem[{Yao et~al.(2023{\natexlab{b}})Yao, Zhao, Yu, Du, Shafran, Narasimhan, and Cao}]{yao2023react}
Shunyu Yao, Jeffrey Zhao, Dian Yu, Nan Du, Izhak Shafran, Karthik Narasimhan, and Yuan Cao. 2023{\natexlab{b}}.
\newblock React: Synergizing reasoning and acting in language models.
\newblock In \emph{International Conference on Learning Representations (ICLR)}.

\bibitem[{Zacks et~al.(2001)Zacks, Tversky, and Iyer}]{zacks2001perceiving}
Jeffrey~M Zacks, Barbara Tversky, and Gowri Iyer. 2001.
\newblock Perceiving, remembering, and communicating structure in events.
\newblock \emph{Journal of experimental psychology: General}, 130(1):29.

\bibitem[{Zeng et~al.(2023)Zeng, Liu, Lu, Wang, Liu, Dong, and Tang}]{zeng2023agenttuning}
Aohan Zeng, Mingdao Liu, Rui Lu, Bowen Wang, Xiao Liu, Yuxiao Dong, and Jie Tang. 2023.
\newblock Agenttuning: Enabling generalized agent abilities for llms.
\newblock \emph{arXiv preprint arXiv:2310.12823}.

\bibitem[{Zhang et~al.(2024)Zhang, Bo, Ma, Li, Chen, Dai, Zhu, Dong, and Wen}]{zhang2024survey}
Zeyu Zhang, Xiaohe Bo, Chen Ma, Rui Li, Xu~Chen, Quanyu Dai, Jieming Zhu, Zhenhua Dong, and Ji-Rong Wen. 2024.
\newblock A survey on the memory mechanism of large language model based agents.
\newblock \emph{arXiv preprint arXiv:2404.13501}.

\bibitem[{Zhao et~al.(2023)Zhao, Zhou, Li, Tang, Wang, Hou, Min, Zhang, Zhang, Dong et~al.}]{zhao2023survey}
Wayne~Xin Zhao, Kun Zhou, Junyi Li, Tianyi Tang, Xiaolei Wang, Yupeng Hou, Yingqian Min, Beichen Zhang, Junjie Zhang, Zican Dong, and 1 others. 2023.
\newblock A survey of large language models.
\newblock \emph{arXiv preprint arXiv:2303.18223}.

\bibitem[{Zheng et~al.(2024)Zheng, Zhang, Zhang, Ye, Luo, Feng, and Ma}]{zheng2024llamafactory}
Yaowei Zheng, Richong Zhang, Junhao Zhang, Yanhan Ye, Zheyan Luo, Zhangchi Feng, and Yongqiang Ma. 2024.
\newblock Llamafactory: Unified efficient fine-tuning of 100+ language models.
\newblock \emph{arXiv preprint arXiv:2403.13372}.

\end{thebibliography}

\clearpage

\appendix

\section{Relevant Prompt}

\label{sec:prompt}

\begin{tcolorbox}[title=Prompt for ALFWorld 3-level hierarchical Plan Collection, width=\linewidth, breakable]
I'll show you  a task in a household environment, along with the sequence of observations and actions an agent took to accomplish it. Based on these observations and actions, I'd like you to create a 3-level hierarchical plan with three levels of detail.\\
\\
The generated plans should be written in the following format:\\
<plan 1>\\
Step 1: ...\\
Step 2: ...\\
...\\
</plan 1>\\
<plan 2>\\
Step 1: ...\\
Step 2: ...\\
...\\
</plan 2>\\
<plan 3>\\
Step 1: ...\\
Step 2: ...\\
...\\
</plan 3>\\
\\
\#\# Notes:\\
\hspace*{1em}- The key difference between plans is only their level of detail, they should all describe the same overall approach\\
\hspace*{1em}- Each successive plan should break down the steps of the previous plan into more detailed substeps\\
\hspace*{1em}- The number of steps should increase with each level: plan 1 has the fewest steps, plan 3 has the most steps\\
\hspace*{1em}- Plan 1 should provide a high-level overview with general steps\\
\hspace*{1em}- Plan 2 should expand on plan 1 by breaking down each high-level step into more specific substeps\\
\hspace*{1em}- Plan 3 should be the most detailed version, breaking down plan 2's steps further, but is still a plan, not executable actions\\
\hspace*{1em}- When creating plan 3, you can reference the following action formats but do not need to refer to specific object\\
\hspace*{2em} 1. go to {{recep}}\\
\hspace*{2em} 2. task {{obj}} from {{recep}}\\
\hspace*{2em} 3. put {{obj}} in/on {{recep}}\\
\hspace*{2em} 4. open {{recep}}\\
\hspace*{2em} 5. close {{recep}}\\
\hspace*{2em} 6. toggle {{obj}} {{recep}}\\
\hspace*{2em} 7. clean {{obj}} with {{recep}}\\
\hspace*{2em} 8. heat {{obj}} with {{recep}}\\
\hspace*{2em} 9. cool {{obj}} with {{recep}}\\
\hspace*{1em}- Ensure all plans describe the same overall approach, just at different levels of detail\\
\hspace*{1em}- Make sure each step is clear, concise, and focused on the "what" rather than the exact "how"\\
\hspace*{1em}- No explanation is needed\\
\#\# Example of a similar task with plans:\\
\{example\_plan\}\\
\\
\# Task: \\
\{task\}\\
\\
\# Interaction sequence:\\
\{interaction\_sequence\}\\
\\
\# Plan:
\end{tcolorbox}

\begin{tcolorbox}[title=Prompt for SciWorld 3-level hierarchical Plan Collection, width=\linewidth, breakable]
I'll show you a task in a scientific environment, along with the sequence of observations and actions an agent took to accomplish it. Based on these observations and actions, I'd like you to create a 3-level hierarchical plan with 3 levels of detail.\\
\\
The generated plans should be written in the following format:\\
<plan 1>\\
Step 1: ...\\
Step 2: ...\\
...\\
</plan 1>\\
<plan 2>\\
Step 1: ...\\
Step 2: ...\\
...\\
</plan 2>\\
<plan 3>\\
Step 1: ...\\
Step 2: ...\\
...\\
</plan 3>\\
\\
\#\# Notes:\\
\hspace*{1em}- The key difference between plans is only their level of detail, they should all describe the same overall approach\\
\hspace*{1em}- Each successive plan should break down the steps of the previous plan into more detailed substeps\\
\hspace*{1em}- The number of steps should increase with each level: plan 1 has the fewest steps, plan 3 has the most steps\\
\hspace*{1em}- Plan 1 should provide a high-level overview with general steps\\
\hspace*{1em}- Plan 2 should expand on plan 1 by breaking down each high-level step into more specific substeps\\
\hspace*{1em}- Plan 3 should be the most detailed version, breaking down plan 2's steps further, but is still a plan, not executable actions\\
\hspace*{1em}- When creating plan 3, you can reference the following action formats but do not need to refer to specific object\\
\hspace*{2em}1.open OBJ: open a container\\
\hspace*{2em}2.close OBJ: close a container\\
\hspace*{2em}3 activate OBJ: activate a device\\
\hspace*{2em}4.deactivate OBJ: deactivate a device\\
\hspace*{2em}5.connect OBJ to OBJ: connect electrical components\\
\hspace*{2em}6.disconnect OBJ: disconnect electrical components\\
\hspace*{2em}7.use OBJ [on OBJ]: use a device/item\\
\hspace*{2em}8.look around: describe the current room\\
\hspace*{2em}9.examine OBJ: describe an object in detail\\
\hspace*{2em}10.look at OBJ: describe a container's contents\\
\hspace*{2em}11.read OBJ: read a note or book\\
\hspace*{2em}12.move OBJ to OBJ: move an object to a container\\
\hspace*{2em}13.pick up OBJ: move an object to the inventory\\
\hspace*{2em}14.pour OBJ into OBJ: pour a liquid into a container\\
\hspace*{2em}15.mix OBJ: chemically mix a container\\
\hspace*{2em}16.teleport to LOC: teleport to a specific room\\
\hspace*{2em}17.focus on OBJ: signal intent on a task object\\
\hspace*{2em}18.wait: task no action for 10 steps\\
\hspace*{2em}19.wait1: task no action for a step\\
\hspace*{1em}- Ensure all plans describe the same overall approach, just at different levels of detail\\
\hspace*{1em}- Make sure each step is clear, concise, and focused on the "what" rather than the exact "how"\\
\hspace*{1em}- No explanation is needed\\
\#\# Action must be in the above format, otherwise it will be invalid. \\
\#\# The condition of action is placed before 'Action: ', for example:\\
\hspace*{2em} step 1: If the blue light bulb is on (metal pot conducts electricity), move the metal pot to the blue box.\\
\hspace*{2em}- Action: move metal pot to blue box\\
\#\# The OBJ in the action must not be specific objects not in Task, for example:\\
\hspace*{2em}- Wrong: Action: pick up the red apple\\
\hspace*{2em}- Right: Action: pick up apple\\

\#\# Example of a similar task with plans:\\
\{example\_plan\}\\
\\
\# Task: \\
You are a helpful assistant to do some scientific experiment in an environment. In the environment, there are several rooms: kitchen, foundry, workshop, bathroom, outside, living room, bedroom, greenhouse, art studio, hallway.\\
\{task\}\\
\\
\# Interaction sequence:\\
\{interaction\_sequence\}\\
\\
\# Plan:
\end{tcolorbox}

\begin{tcolorbox}[title=Prompt for AlfWorld self-Adaptive Hierarchical Plan Generation, width=\linewidth, breakable]
Please generate hierarchical plans for a household task based on its difficulty level:\\
<task>\\
\{task\}\\
</task>\\
\\
The number of plans and their levels of detail should adjust according to the difficulty of the task:\\
\hspace*{2em}- Easy tasks should have 1 level plan (high-level overview).\\
\hspace*{2em}- Medium tasks should have 2 level plans (overview and a bit more detail).\\
\hspace*{2em}- Difficult tasks should have 3 level plans (overview, intermediate detail, and highly detailed breakdown).\\
\\
The generated plans should be written in the following format:\\
<plan 1>\\
Step 1: ...\\
Step 2: ...\\
...\\
</plan 1>\\
...\\
<plan N>\\
Step 1: ...\\    
Step 2: ...\\
...\\
</plan N>\\
Where N is the number of plan levels based on task difficulty.\\
\\
\# Your plan for this task:
\end{tcolorbox}

\begin{tcolorbox}[title=Prompt for SciWorld self-Adaptive Hierarchical Plan Generation, width=\linewidth, breakable]
Please generate hierarchical plans for a scientific task in a scientific environment based on its difficulty level.
The task difficulty can be inferred from the environment description and the task itself.\\
\\
<task>\\
You are a helpful assistant to do some scientific experiment in an environment.\\
In the environment, there are several rooms: kitchen, foundry, workshop, bathroom, outside, living room, bedroom, greenhouse, art studio, hallway.\\
\{task\}\\
</task>\\
\\
The generated plans should be written in the following format:\\
<plan 1>\\
Step 1: ...\\
Step 2: ...\\
...\\
</plan 1>\\
...\\
<plan N>\\
Step 1: ...\\    
Step 2: ...\\
...\\
</plan N>\\
Where N is the number of plan levels based on task difficulty.\\
\\
\#\# Notes:
The number of plans and their levels of detail should adjust according to the difficulty of the task:\\
\hspace*{1em}- Easy tasks should have 1 level plan (high-level overview).\\
\hspace*{1em}- Medium tasks should have 2 level plans (overview and a bit more detail).\\
\hspace*{1em}- Difficult tasks should have 3 level plans (overview, intermediate detail, and highly detailed breakdown).\\
\hspace*{1em} When creating plan 3, you can reference the following action formats but do not need to refer to specific object\\
\hspace*{2em}1.open OBJ: open a container\\
\hspace*{2em}2.close OBJ: close a container\\
\hspace*{2em}3 activate OBJ: activate a device\\
\hspace*{2em}4.deactivate OBJ: deactivate a device\\
\hspace*{2em}5.connect OBJ to OBJ: connect electrical components\\
\hspace*{2em}6.disconnect OBJ: disconnect electrical components\\
\hspace*{2em}7.use OBJ [on OBJ]: use a device/item\\
\hspace*{2em}8.look around: describe the current room\\
\hspace*{2em}9.examine OBJ: describe an object in detail\\
\hspace*{2em}10.look at OBJ: describe a container's contents\\
\hspace*{2em}11.read OBJ: read a note or book\\
\hspace*{2em}12.move OBJ to OBJ: move an object to a container\\
\hspace*{2em}13.pick up OBJ: move an object to the inventory\\
\hspace*{2em}14.pour OBJ into OBJ: pour a liquid into a container\\
\hspace*{2em}15.mix OBJ: chemically mix a container\\
\hspace*{2em}16.teleport to LOC: teleport to a specific room\\
\hspace*{2em}17.focus on OBJ: signal intent on a task object\\
\hspace*{2em}18.wait: task no action for 10 steps\\
\hspace*{2em}19.wait1: task no action for a step\\
\\
\# Your plan for this task:\\
\end{tcolorbox}

\section{Case Study}
\begin{tcolorbox}[title=A 3-level hierarchical plan for ALFWorld, width=\linewidth, breakable]
Task: \\
You are in the middle of a room. Looking quickly around you, you see a cabinet 4, a cabinet 3, a cabinet 2, a cabinet 1, a coffeemachine 1, a countertop 1, a diningtable 3, a diningtable 2, a diningtable 1, a drawer 1, a fridge 1, a garbagecan 1, a microwave 1, a sidetable 1, a sinkbasin 1, a stoveburner 4, a stoveburner 3, a stoveburner 2, a stoveburner 1, and a toaster 1.
\\Your task is to: find some apple and put it in sidetable.\\
\\
Plan:\\
<plan 1>\\
Step 1: Find the apple.\\
Step 2: Take the apple.\\
Step 3: Put the apple in/on the sidetable.\\
</plan 1>\\
<plan 2>\\
step 1: Search the room for the apple.\\ 
\hspace*{1em}(likely locations: fridge, dining tables, sidetable, countertop, sinkbasin, garbagecan). \\
Step 2: Once the apple is found, pick it up.\\
Step 3: Move to the sidetable.\\
Step 4: Place the apple in/on the sidetable.\\
</plan 2>\\
<plan 3>\\
Step 1: Examine potential locations for the apple. Start by checking the fridge.\\ 
Step 2: Open the fridge if necessary and search inside for the apple.\\
Step 3: If the apple is not found, check the dining tables one by one.\\
Step 4: If the apple is still not found, check the sidetable.\\
Step 5: If the apple is still not found, check the countertop, then the sinkbasin.\\
Step 6: Finally, check the garbagecan.\\
Step 7: Once the apple is found, take it from its location.\\
Step 8: Move to the sidetable.\\
Step 9: Observe the sidetable for available space.\\
Step 10: Place the apple in/on the sidetable.\\
</plan 3>\\
\end{tcolorbox}

\begin{tcolorbox}[title=A 3-level hierarchical plan for ScienceWorld, width=\linewidth, breakable]
Task: \\
You are a helpful assistant to do some scientific experiment in an environment.
In the environment, there are several rooms: kitchen, foundry, workshop, bathroom, outside, living room, bedroom, greenhouse, art studio, hallway.\\
\\
Your task is to use chemistry to create green paint. When you are done, focus on the green paint.\\
\\
Plan:\\
<plan 1>\\
Step 1: Teleport to the art studio.\\
Step 2: Look around to find blue and yellow paint.\\
Step 3: Mix the blue and yellow paints to create green paint.\\
Step 4: Focus on the green paint.\\
</plan 1>\\
<plan 2>\\
Step 1: Teleport to the art studio.\\
Step 2: Look around and find the cupboard with paint supplies.\\
Step 3: Identify the blue and yellow paints.\\
Step 4: Pour blue paint into an empty container.\\
Step 5: Pour yellow paint into the same container.\\
Step 6: Mix the paints to create green paint.\\
Step 7: Focus on the green paint.\\
</plan 2>\\
<plan 3>\\
Step 1: Teleport to the art studio.\\
\hspace*{1em}- Possible Action: teleport to art studio\\
Step 2: Look around the art studio and locate the cupboard containing yellow, blue, and red paints.\\
\hspace*{1em}- Possible Action: look around\\
Step 3: Identify the blue and yellow paints as the necessary ingredients to make green paint.\\
\hspace*{1em}- Possible Action: examine blue paint\\
\hspace*{1em}- Possible Action: examine yellow paint\\
Step 4: Pour the contents of the blue paint cup into an empty glass cup.\\
\hspace*{1em}- Possible Action: pour blue paint into cup containing nothing\\
Step 5: Pour the contents of the yellow paint cup into the same glass cup that contains blue paint.\\
\hspace*{1em}- Possible Action: pour yellow paint into cup containing blue paint\\
Step 6: Mix the blue and yellow paints together in the glass cup to create green paint.\\
\hspace*{1em}- Possible Action: mix blue and yellow paint\\
Step 7: Focus on the green paint to complete the task.\\
\hspace*{1em}- Possible Action: focus on green paint\\
</plan 3>\\
\end{tcolorbox}

\section{Data Statistics}
\label{sec: data}
As shown in Table~\ref{tab:sta}, we construct hierarchical plans based on the correct trajectories in the training set of two datasets, and then test them separately on Seen and Unseen.
\begin{table}[htbp]
  \centering
  \caption{Statistical information of the two datasets.}
  \resizebox{\linewidth}{!}{
    \begin{tabular}{cccc}
\hline
\hline
\bf{Datasets} & \bf{Train} & \bf{Valid(Seen)} & \bf{Test(UnSeen)}\\
\hline
ALFWorld & 1483 & 194 & 211  \bigstrut[t]\\
ScienceWorld &  3553 & 140 & 134  \\
\hline
\hline
\end{tabular}%
}
  \label{tab:sta}
\end{table}%

\begin{figure}[t!]
    \vspace{-10.5cm}
    \begin{minipage}[b]{0.45\textwidth}
        \centering
        \includegraphics[width=\linewidth]{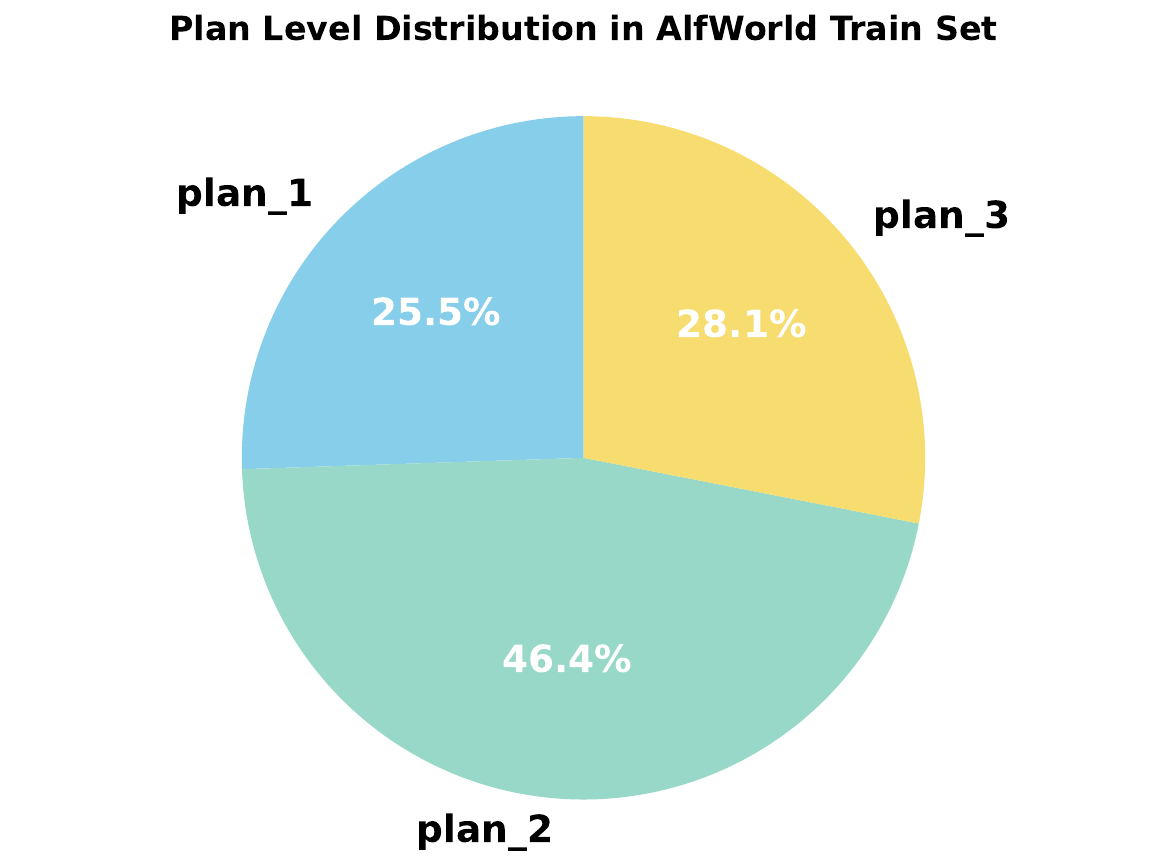}
        \caption{The distribution of the optimal number of levels for the hierarchical plans corresponding to each task in the training sets of ALFWorld.}
        \label{fig:alf}
    \end{minipage}
    \hfill
    \begin{minipage}[b]{0.45\textwidth}
        \centering
        \includegraphics[width=\linewidth]{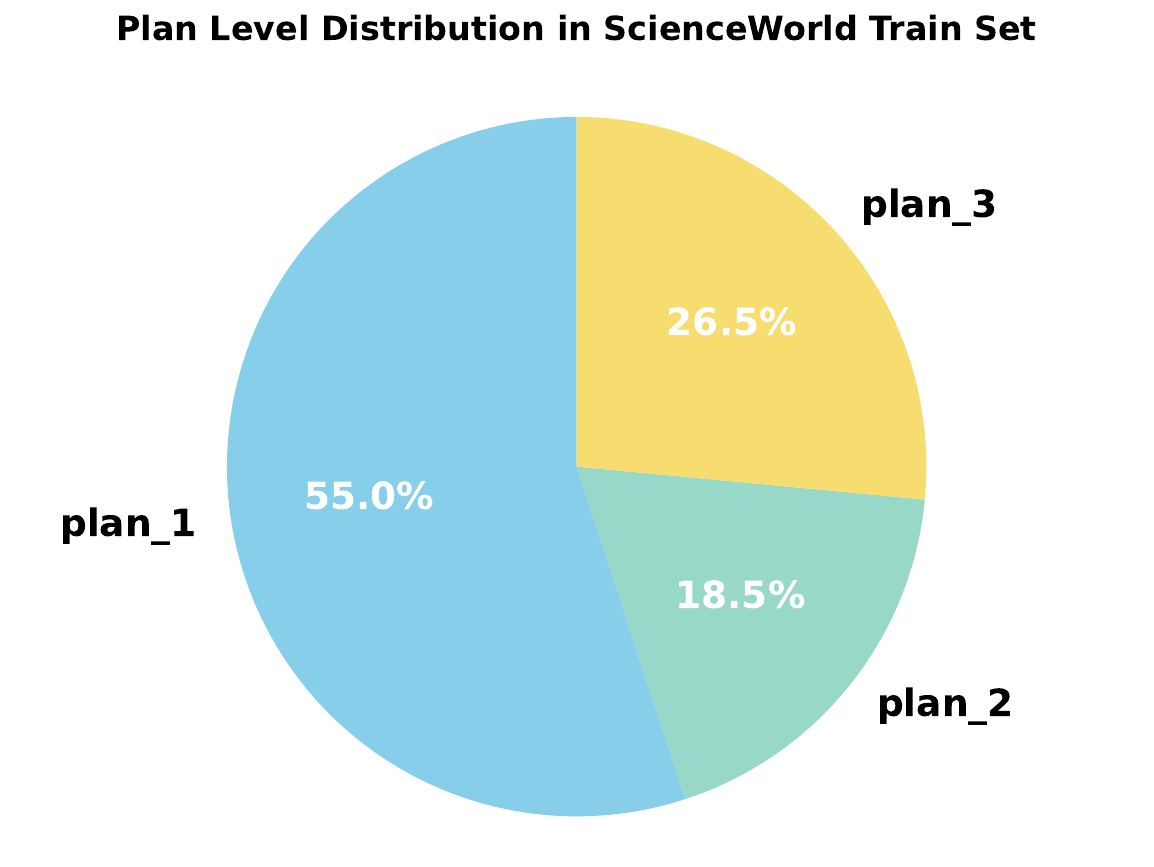}
        \caption{The distribution of the optimal number of layers for the hierarchical plans corresponding to each task in the training sets of ScienceWorld.}
        \label{fig:sci}
    \end{minipage}
\end{figure}

In the first stage of our training, the value of best levels of the synthesized and selected hierarchical plans from the training sets of the two datasets is not exactly the same, as shown in Figure~\ref{fig:alf} and Figure~\ref{fig:sci}.




\end{document}